# SwishReLU: A Unified Approach to Activation Functions for Enhanced Deep Neural Networks Performance

**Jamshaid Ul Rahman[1,2,*], Rubiqa Zulfiqar [2], Asad Khan[3] , Nimra[2]**

[1]School of Mathematical Sciences, Jiangsu University 301 Xuefu road, Zhenjiang 212013, China.
[2]Abdus Salam School of Mathematical Sciences, Govt College University, Lahore 54600, Pakistan.
[3]Metaverse Research Institute, School of Computer Science and Cyber Engineering, Guangzhou University, Guangzhou 510006, China.

Corresponding author: [*]jamshaidrahman@gmail.com

## Abstract

ReLU, a commonly used activation function in deep neural networks, is prone to the issue of "Dying ReLU". Several enhanced versions, such as ELU, SeLU, and Swish, have been introduced and are considered to be less commonly utilized. However, replacing ReLU can be somewhat challenging due to its inconsistent advantages. While Swish offers a smoother transition similar to ReLU, its utilization generally incurs a greater computational burden compared to ReLU. This paper proposes SwishReLU, a novel activation function combining elements of ReLU and Swish. Our findings reveal that SwishReLU outperforms ReLU in performance with a lower computational cost than Swish. This paper undertakes an examination and comparison of different types of ReLU variants with SwishReLU. Specifically, we compare ELU and SeLU along with Tanh on three datasets: CIFAR-10, CIFAR-100 and MNIST. Notably, applying SwishReLU in the VGG16 model described in Algorithm 2 yields a 6% accuracy improvement on the CIFAR-10 dataset.

## 1. Introduction

Deep neural networks [1] have made remarkable advancements in various research domains, tackling a wide range of challenges such as object detection [2], scene classification [3], semantic segmentation [4-5], skin disease diagnosis [6], anomaly detection [7], and many more. These sophisticated computational structures, recognized for their multi-layered architecture, have significantly impacted various fields, ranging from artificial intelligence to academic and industrial applications. Their adaptability and learning capabilities have contributed to significant milestones, particularly in areas such as computer vision [8], natural language processing [9], and biomedical research [10].

In recent years, there have been substantial enhancements in the representation and recognition performance of neural networks employing deep architectures [11-12]. The structural framework of Deep Neural Networks (DNNs) typically includes an input layer, several hidden layers, and an output layer. Within each layer, nodes work collaboratively to process information, iteratively enhancing it as it traverses through the network. Assume the input features are represented as a vector $x$. Then for each hidden layer $l$ with weights $W^{(l)}$ and biases $b^{(l)}$, the outputs $a^{(l)}$ and pre-activation $z^{(l)}$ can be calculated as follows (described in equation (1.1) and (1.2)):

$$z^{(l)} = W^{(l)}.a^{(l-1)} + b^{(l)} \quad (1.1)$$

$$a^{(l)} = \sigma(z^{(l)}) \quad (1.2)$$

Here, $\sigma(.)$ is the activation function.

And for the output layer L with weights $W^{(L)}$ and biases $b^{(L)}$ as described in equation (1.3) and (1.4):

$$z^{(L)} = W^{(L)}.a^{(L-1)} + b^{(L)} \quad (1.3)$$

$$\hat{y} = \sigma(z^{(L)}) \quad (1.4)$$

Here, $y^{(l)}$ is the predicted output.

The performance of deep neural networks relies on various factors, with the evolution of the activation function standing out as one of the most crucial elements [13]. The activation function plays a pivotal role in shaping the learning and decision-making processes within the network [14], influencing its ability to capture intricate patterns and representations in data. As research continues to advance in this domain, understanding and refining activation functions become integral to unlocking the full potential of deep neural networks in diverse applications. The development of activation functions plays a vital role in investigating the performance of works [15]. Consequently, an increasing amount of effort has been focused on the research and study of activation functions [16-19].

Activation functions play a crucial role in artificial neural networks [20]. Their primary function is to transform an input signal into an output signal, which is subsequently passed as input to the succeeding layer within the neural network. Within an artificial neural network [21], the process involves calculating the sum of products of inputs and their corresponding weights. Subsequently, an activation function is applied to this sum to obtain the output of that particular layer, which is then transmitted as input to the next layer. It involves a structure with hidden layers, usually there is more than one layer in the network [22].

The prediction accuracy of a neural network is determined by the choice of activation function. In the absence of an activation function, a neural network operates akin to a linear regression model [23]. In this scenario, the predicted output essentially replicates the input provided, limiting the network's capacity to capture intricate patterns and relationships within the data. To overcome this limitation and introduce the necessary complexity for effective learning, non-linear activation functions [24] are commonly employed.

A linear activation function exhibits a linear boundary, limiting the network's adaptability to linear changes in the input. In real world scenarios, errors often possess non-linear behavior. Given the neural network's [25] capacity to learn from erroneous data, non-linear activation functions are favored over linear activation functions in order to capture and model these non-linear relationships.

An essential characteristic of an activation function is its differentiability, enabling the implementation of backpropagation optimization strategy [26-27]. The iterative process of backpropagation, a fundamental component of deep learning [28-29], refines the connections between neurons by minimizing a predefined loss function. This differentiation is crucial for computing errors or losses concerning weights, facilitating the optimization of weights through techniques like Gradient Descent or other optimization methods to minimize errors [30-32]. By minimizing errors iteratively, the neural network refines its parameters, enhancing its ability to accurately capture complex patterns and relationships in the data during the training process.

Initially, in the early stages of neural network development, shallow networks utilized sigmoid or tanh nonlinearities as activations. However, challenges arose as network depth increased, making it increasingly difficult [33] to train deeper networks with these functions [34]. The limitations became apparent, prompting a shift towards the adoption of alternative activation functions, such as Rectified Linear Unit (ReLU) and its variants.

Nair and Hinton's [35] introduction of ReLU defined in equation (1.5), marked a significant development in addressing the challenges associated with saturated activation functions like sigmoid and tanh. To overcome the Vanishing Gradient problem [36] and to enhance the performance, these saturated functions were replaced with their non-saturated counterparts, such as ReLU and ELU [37]. Deep networks incorporating ReLU are more readily optimized, making ReLU the default activation function widely adopted in the deep learning community [38]. But Unfortunately, ReLU has a significant problem known as "Dying ReLU" [39]. In cases where a neuron receives negative inputs, causing it to remain inactive, there is a risk of omitting weight updates during the backpropagation process, hindering the learning process and potentially impeding the overall performance of the neural network.

$$f(x) = \max\{0, x\} \qquad (1.5)$$

Various alternatives to traditional activation functions have been proposed, and experimental results have showcased the superior performance of Swish [40], defined in equation (1.6), when compared to ReLU. But There are also some problems with Swish. Firstly, the Swish function can be computationally more complex compared to the ReLU. Its complexity may result in longer calculation times, particularly when dealing with large datasets [41]. Moreover, the Swish function can exhibit instability when applied to very deep neural networks, resulting in slower training or potentially leading to divergent behavior. This paper introduces a novel activation function named SwishReLU, aiming to synergize elements from two well-established activation functions, ReLU and Swish. By combining features from both, SwishReLU seeks to leverage the strengths of

ReLU's simplicity and Swish's effectiveness, presenting a potential enhancement in performance and versatility for neural network architectures.

$$f(x) = \frac{x}{1+e^{-x}} \quad (1.6)$$

Ensuring continuous differentiability stands as a pivotal characteristic for activation functions, particularly in facilitating the efficacy of gradient-based optimization methods. The capacity to compute gradients seamlessly throughout the function's domain is fundamental for the successful application of optimization algorithms such as gradient descent [42]. In contrast to ReLU, which lacks continuous differentiability, our newly introduced SwishReLU excels in this aspect. This property enhances its alignment with gradient-based optimization, rendering SwishReLU a promising choice for achieving efficient and effective training in neural networks. The MNIST [43], CIFAR-10, and CIFAR-100 [44] datasets are widely recognized as benchmark datasets for testing the performance of algorithms, particularly in the area of image classification [45-46]. By serving as standardized benchmarks, these datasets enable the comprehensive evaluation of the robustness, accuracy, and generalization capabilities of various machine learning and deep learning models on complex visual recognition tasks.

Through extensive experiments, our findings demonstrate that SwishReLU consistently surpasses or equal to ReLU when applied to deep networks across various challenging domains, including tasks like image classification. Notably, in the case of the VGG16 model, SwishReLU demonstrated a substantial improvement, achieving a 6% relative performance gain compared to ReLU for the classification of CIFAR-10 dataset [47]. Ongoing exploration in this field continues to refine and innovate activation functions for enhanced neural network performance.

2. **SwishReLU**

We proposed a new activation function, which we call SwishReLU as defined in equation 2.1.

$$f(x) = \begin{cases} \frac{x}{1+e^{-x}} & if\ x < 0 \\ x & if\ x \geq 0 \end{cases} \quad (2.1)$$

SwishReLU operates similarly to ReLU for non-negative values, maintaining comparable characteristics. However, for negative values, it adopts features reminiscent of Swish, addressing the issue of Dying ReLU. This dual functionality allows SwishReLU to effectively handle both positive and negative input values, contributing to improved performance and mitigating challenges associated with the traditional ReLU activation function. Moreover, it incurs lower computational costs compared to Swish. As ReLU is not zero centered so it can contribute to issues such as vanishing gradients during training, particularly in deep neural networks. SwishReLU, on the other hand, is designed to be zero-centered, which can help mitigate problems associated with gradient vanishing.

Similar to both ReLU and Swish, SwishReLU is unbounded above and bounded below. SwishReLU is a function that exhibits differentiability, meaning it can be smoothly differentiated throughout its range. The dichotomous behavior of SwishReLU enables it to blend the advantageous aspects of ReLU and Swish. This versatility allows SwishReLU to effectively capture both linear and non-linear patterns in data, making it a valuable choice in various contexts. In Figure 1, a visual representation is presented, showcasing plots that depict the behavior of different activation functions. Among these functions is SwishReLU, a key inclusion in the comparison. This visual depiction allows for a detailed examination of the distinctive characteristics exhibited by various activation functions.

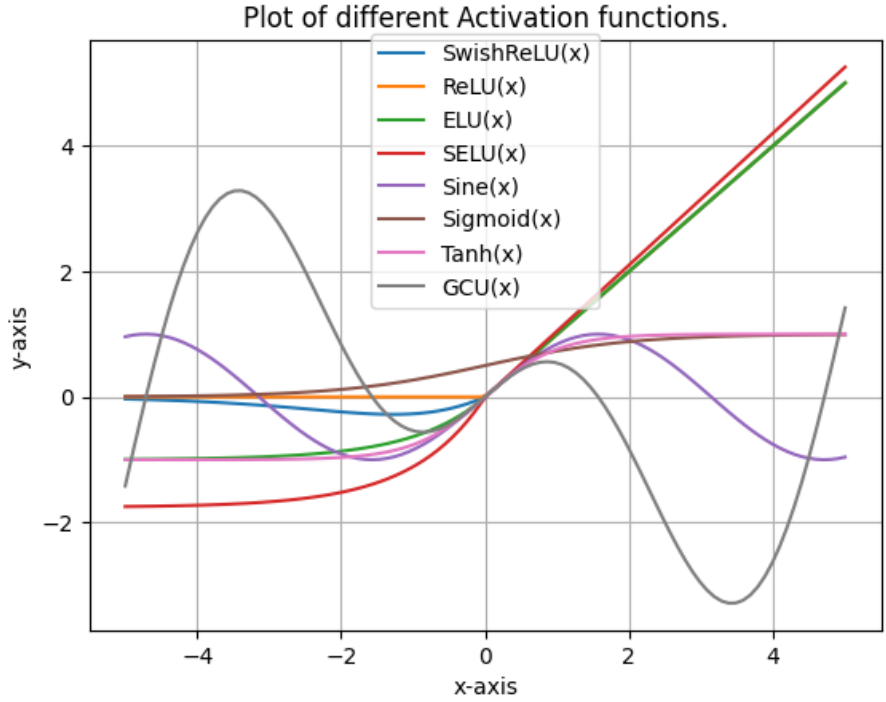

Figure 1: Plot of Different Activation functions

**3. Data used and SwishReLU infused Architectures**

**3.1 Data**

Initiating a project with an excellent database is always a commendable beginning. In this research, our approach involves adopting the MNIST, CIFAR-10 and CIFAR-100 datasets. By choosing these databases, we aim to lay a strong foundation for meaningful insights and findings in the subsequent phases of our research.

The MNIST [48] dataset, short for the Modified National Institute of Standards and Technology comprising a collection of handwritten digit images as shown in Figure 2, it serves as a fundamental benchmark for various machine learning tasks. It consists of 60,000 training images set and 10,000 for testing set, all drawn from the same distribution. These images depict black and white digits that are size-normalized and centered within fixed-size frames of 28x28 pixels.

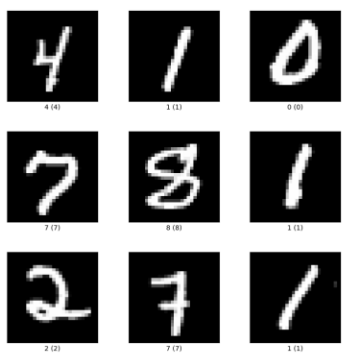

Figure 2: Handwritten digit images from the MNIST dataset.

The CIFAR-10 dataset, which stands for the Canadian Institute for Advanced Research, comprises a collection of images widely employed for training machine learning and computer vision algorithms. It contains a total of 60,000 images, and these images are organized into ten distinct classes, namely Airplane, Automobile, Bird, Cat, Deer, Dog, Frog, Horse, Ship, Truck. The composition of the CIFAR-10 dataset, illustrated in Figure 3. Every image within the CIFAR-10 dataset is sized at 32x32 pixels and is structured with three channels corresponding to the Red, Green, and Blue (RGB) color components. As a result, the shape of each data frame representing an image is (32, 32, 3). For the dataset distribution, there are a total of 50,000 images specifically designated for training purposes, facilitating the learning process of deep learning models. Additionally, there are 10,000 images set aside for testing, allowing for an effective evaluation of the trained models on new, unseen data.

CIFAR-100 serves as an extension to the CIFAR-10 dataset, originating from the Canadian Institute for Advanced Research. This expanded dataset encompasses a collection of 60,000 color images, each with dimensions of 32x32 pixels. In contrast to CIFAR-10, CIFAR-100 introduces a more diverse classification challenge by categorizing each image into one of 100 distinct classes. Figure 3 provides a visual representation of the CIFAR-100 dataset. The CIFAR-100 dataset introduces a finer granularity by incorporating 100 distinct classes, offering a more extensive array of objects and concepts compared to CIFAR-10. This increased diversity in classes makes CIFAR-100 a more challenging dataset for machine learning and computer vision tasks in comparison to CIFAR-10. The broader range of objects and concepts challenges algorithms to develop more nuanced and sophisticated recognition capabilities, enhancing their ability to discern finer distinctions among a diverse set of images.

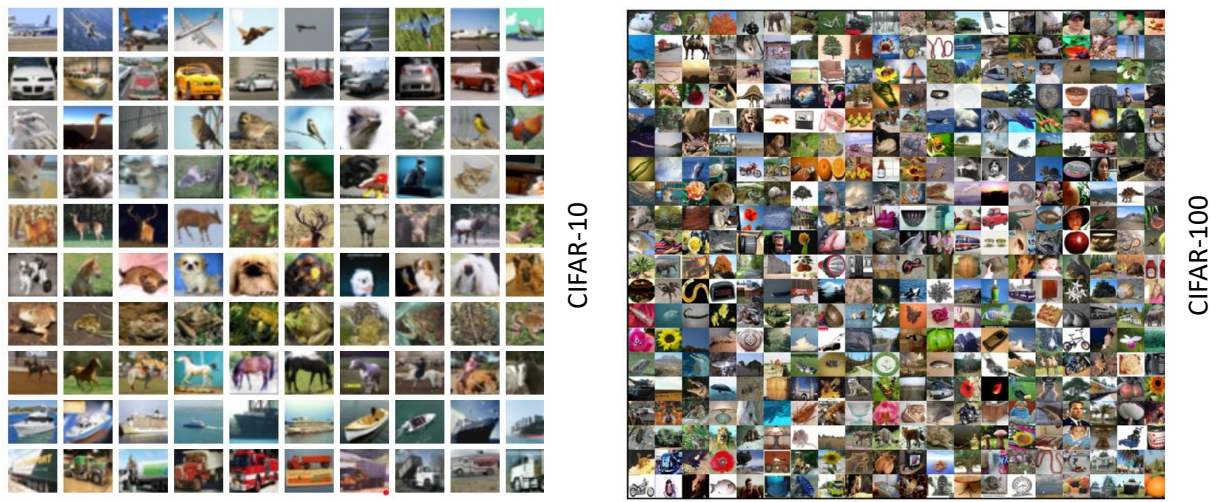

Figure 3: Images of CIFAR-10 and CIFAR-100 dataset

## 3.2 Proposed Model

The following architectures are employed for classifying the images within our datasets.

### 3.2.1 Utilizing SwishReLU in FCNN

Fully connected neural networks (FCNN), also known as dense or multi-layer perceptron (MLPs), are a type of artificial neural network where each neuron in one layer is connected to every neuron in the next layer [49]. In the context of the MNIST dataset, fully connected neural networks have been widely employed for digit classification tasks. The FCNN architecture has 300 neurons in the input layer and 10 neurons in the output layer, and a hidden layer with 100 neurons as shown in Figure 4.

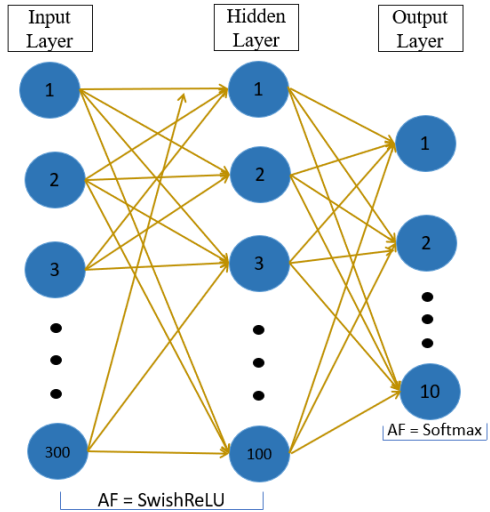

Figure 4: Fully Connected Neural Network with one hidden layer containing 100 neurons.

### 3.2.2 Infusion of SwishReLU in CNN Model

CNNs are widely acknowledged as highly effective architectures for processing and analyzing image data. The CNN architecture is characterized by the presence of three fundamental types of layers: convolutional layers (Conv), pooling layers, and fully-connected (FC) layers. The arrangement and stacking of these layers define the structural composition of a CNN.

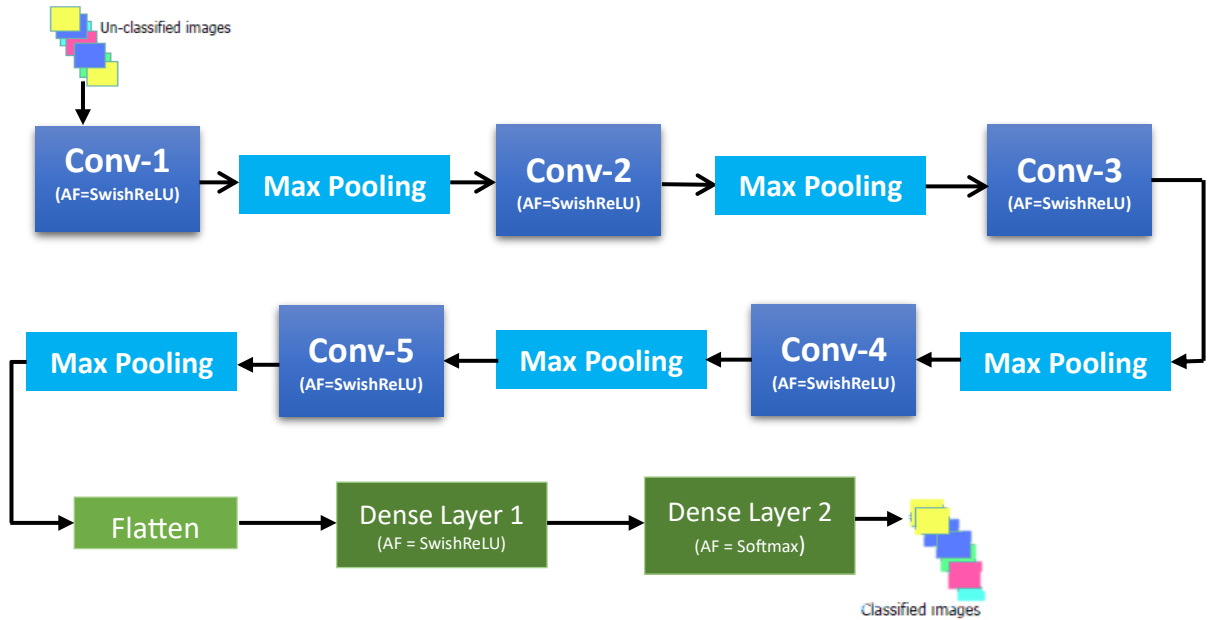

Figure 5: CNN Architecture with five convolutional layers

In the context of this study, Figure 5 offers a comprehensive visual representation of the CNN architecture implemented for the classification tasks involving both the CIFAR-10 and CIFAR-100 datasets. This figure serves as an illustrative guide, showcasing the arrangement and connectivity of the convolutional, pooling, and fully-connected layers within the CNN framework utilized in the research.

### 3.2.3 VGG-16 Model

VGG16 [50], also known as VGGNet, stands as a CNN model characterized by a total of 16 layers. Within this architectural configuration, 13 layers are dedicated to convolutional operations, while the remaining 3 layers are dense layers. Figure 6 serves as a visual representation, illustrating the intricate structure of the VGG16 Model. This graphical depiction effectively communicates the diverse layers, connections, and patterns embedded within the model, providing a comprehensive overview of its architectural intricacies.

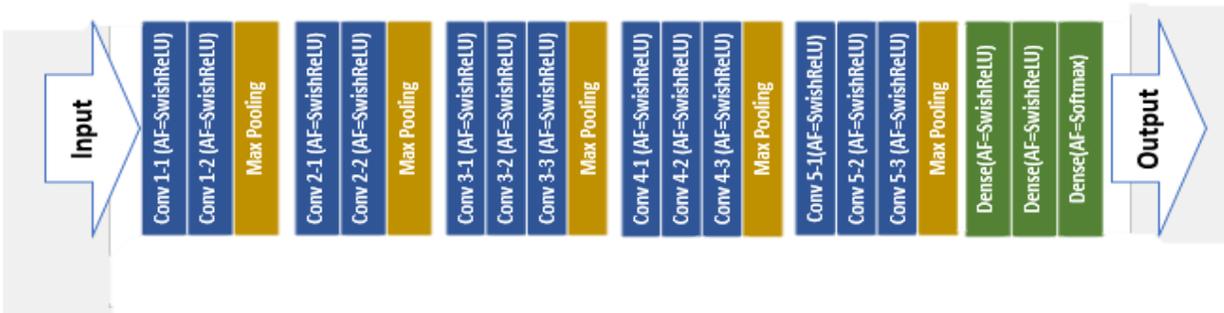

Figure 6: VGG16 Architecture with SwishReLU as an activation function

### 3.2.4 Algorithms

To facilitate a more comprehensive comprehension of our structural configurations, we have explained the CNN and VGG-16 model in Algorithm 1 and Algorithm 2, respectively. These algorithms provide a detailed step-by-step explanation of the complexities within each model, aiming for a clear understanding of the frameworks employed in our study.

*Algorithm 1:* Employing SwishReLU in CNN

| | |
|---|---|
| *Input:* Datasets CIFAR 10 and CIFAR-100<br>*Models:* 5-layer CNN<br>*Optimizer:* Adam<br>*Activation Functions:* SwishReLU, ReLU, ELU, SeLU, Tanh<br>*Output:* Classification of Images | **Procedure**<br>Step 1    LOAD dataset X:<br>            SPLIT dataset into training samples (X-train) and validation samples (X-validation)<br>Step 2    EXTRACT X-train<br>            EXTRACT X-validation<br>Step 3    for the no. of activation-functions do<br>              LOAD X-train<br>              5-layer CNN model<br>              Training Process<br>              Update weight using backpropagation and the optimizer<br>              $W_i = W_i - \alpha \frac{\partial L}{\partial W_i}$<br>              LOAD X-validation<br>              $Y_t$= 5-layer CNN PREDICT (X-test)<br>              GENERATE accuracy metric<br>           end for<br>Step 4    Competency comparison of different model accuracy metric<br>              REPORT results<br><br>End procedure |

*Algorithm 2:* Utilizing SwishReLU in the VGG-16 Model

| | |
|---|---|
| *Input:* Dataset CIFAR 10<br>*Models:* VGG-16 Model<br>*Optimizer:* SGD (learning rate 0.01, momentum = 0.9)<br>*Activation Functions:* SwishReLU<br>*Output:* Classification of Images | **Procedure**<br>Step 1    LOAD dataset X:<br>            SPLIT dataset into training samples (X-train) and validation samples<br>            (X-validation)<br>Step 2    EXTRACT X-train<br>            EXTRACT X-validation<br>Step 3    VGG-16 Model<br>            Set Early stopping Criteria (monitor = val_loss, patience = 5)<br>            for epoch in range (maximum epochs)<br>              Model fitting<br>              Training Process<br>              Update weight using backpropagation and the optimizer<br>              $W_i = W_i - \alpha \frac{\partial L}{\partial W_i}$<br><br>              LOAD X-validation<br>              $Y_t$= VGG-16 Model PREDICT (X-test) |

```
                If early_stopping_criteria_met:
                    Break
                end for
Step 4      Competency comparison of model accuracy
metric
                REPORT results
End procedure
```

## 3.3. Experimental Settings:

We conducted an evaluation of SwishReLU in comparison to ReLU and other activation functions, focusing on challenging datasets. The findings reveal that SwishReLU consistently either matches or surpasses the baseline performance across nearly all tasks. This underscores the efficacy of SwishReLU as an activation function. In our study, we utilized the FCNN and the CNN architecture infused with SwishReLU illustrated in Figure 4 and in Figure 5 respectively. The CNN architectural configuration comprises five convolutional layers, followed by dense layers in its design. In the training phase, both the FCNN and CNN architectures employ Sparse Categorical Cross-entropy loss, which is subsequently optimized using the Adam optimizer. The training duration is set to cover 30 epochs for the FCNN and 50 epochs for the CNN. This signifies the number of times the entire dataset is iterated over by the respective neural network architectures during the training phase. The same SwishReLU infused CNN architecture employed for CIFAR-10 was utilized for CIFAR-100, with the only modification being the replacement of the final softmax layer containing 10 neurons with a new softmax layer comprising 100 neurons. The utilization of this specific model, as visually represented in Figure 5, served as the foundation for our experimental approach. Following that, we proceeded to evaluate the accuracy of the SwishReLU activation function within the context of the VGG16 architecture, applied to both CIFAR-10 and CIFAR-100 dataset. The representation of feature map visualization has been carried out for the CNN and VGG-16 architectures. The visualization of these feature maps serves as a critical process through which a deeper understanding of the learned features and intricate patterns within the network is attained.

## 4. Results and Discussion:

In this section, we provide an in-depth summary of the experimental results obtained through the utilization of the SwishReLU in FCNN for MNIST dataset and SwishReLU-infused CNN architecture for the classification tasks involving both the CIFAR-10 and CIFAR-100 datasets. Additionally, we present a summary of the outcomes derived from employing the VGG-16 model with SwishReLU as an activation function for the classification of the both CIFAR-10 and CIFAR-100 datasets. This comparative evaluation aims to elucidate the performance characteristics of SwishReLU in the context of image classification on these datasets, providing valuable insights into its effectiveness as compared to other activation functions employed in similar scenarios.

The observed improvement in accuracy with SwishReLU can be attributed to its ability to address the issue of "Dying ReLU." SwishReLU diminishes the problem by assigning non-zero values to negative weights during the activation process. This characteristic helps prevent neurons from becoming inactive, allowing for the effective flow of information through the network and contributing to enhanced learning and performance in various tasks.

The assessment of the network encompassed an examination of its performance metrics, with a specific emphasis on accuracy concerning both the training and testing datasets. Furthermore, a detailed analysis of the trajectory of the loss function throughout both the training and testing phases was undertaken. The summarized findings are visually presented in Figures 7 to Figure 14, providing a comprehensive overview of the network's learning dynamics and performance characteristics.

From the Figure 7 and Figure 8, it's apparent that SwishReLU is consistently beating ReLU and other activation functions in terms of accuracy on the MNIST dataset. Figure 7 depicting the Training Accuracy and Loss and Figure 8 shows the Testing Accuracy and Loss of the FCNN architecture on MNIST.

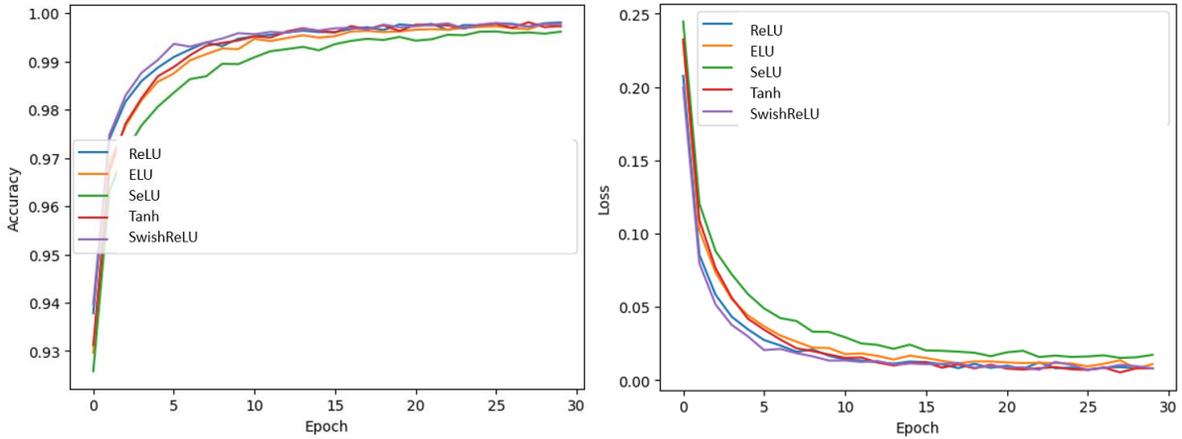

Figure 7: Training Accuracy(left) and Loss(right) in FCNN architecture for MNIST.

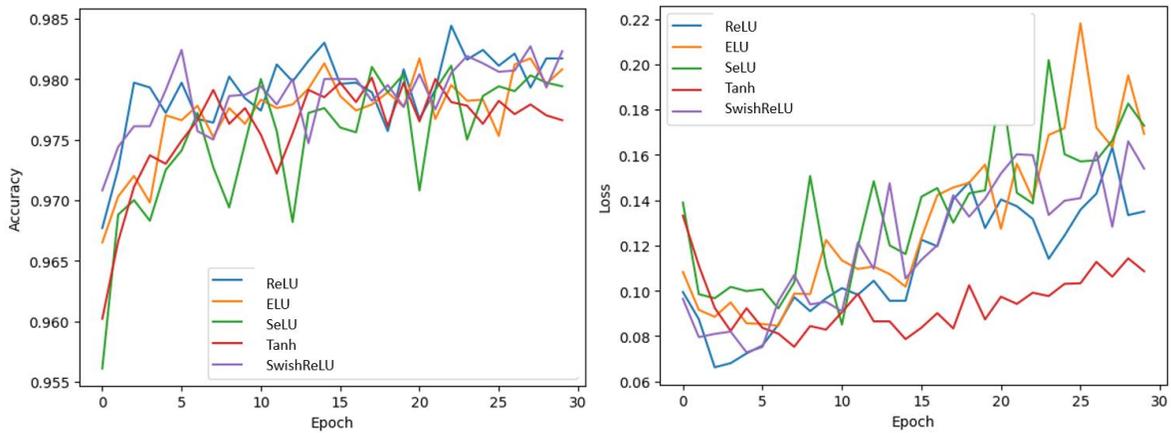

Figure 8: Testing Accuracy(left) and Loss(right) in FCNN architecture for MNIST.

As illustrated in Figures 9 and Figure 10, the accuracy of the SwishReLU activation function exhibits an upward trend with the progression of epochs. Notably, it surpasses the performance of other activation functions, ultimately achieving the highest accuracy on the CIFAR-10 dataset. In Figure 9, the depiction of training accuracy and training loss is presented. Notably, the purple line representing SwishReLU illustrates a noticeable upward trend in accuracy and a simultaneous reduction in loss. Figure 10 shows the testing accuracy and loss of SwishReLU, presenting a comparative view with other activation functions. This visual observation suggests that SwishReLU contributes to an improved accuracy and more effective learning, as indicated by the favorable trajectory in both accuracy and loss metrics.

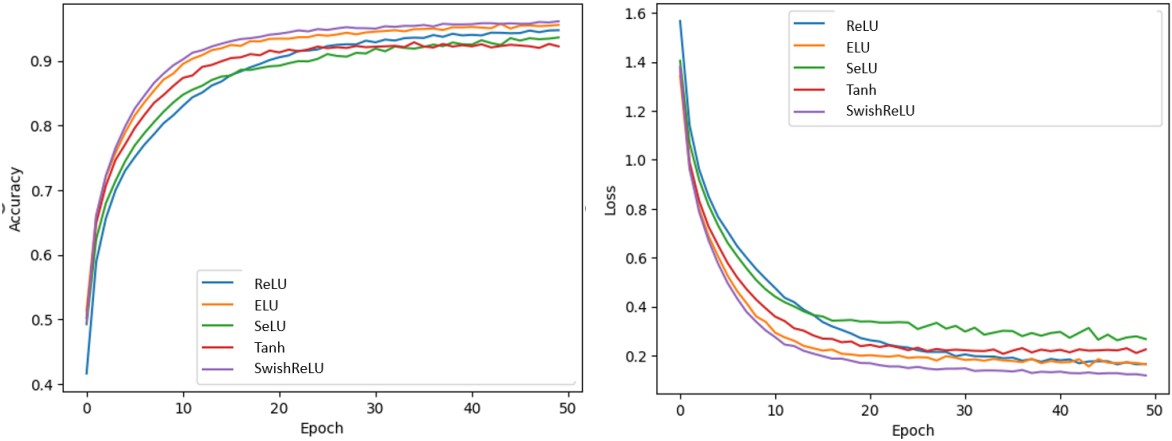

Figure 9: Training data accuracy(left) and loss(right) on the CIFAR-10 dataset.

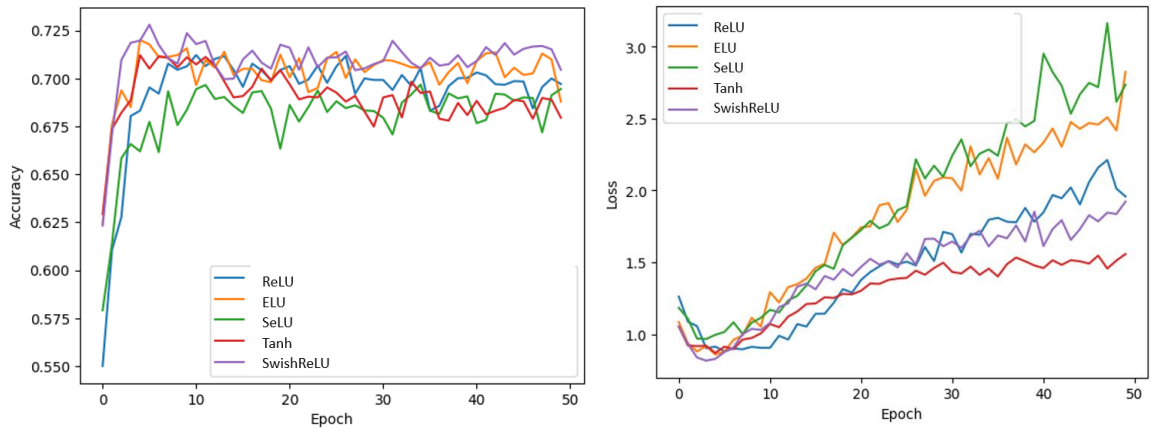

Figure 10: Testing data accuracy(left) and loss(right) on the CIFAR-10 dataset.

The observed consistent pattern seamlessly extends to the CIFAR-100 dataset, demonstrating its enduring presence in both training and testing metrics. This continuity is visually depicted in Figures 11 and Figure 12, where Figure 11 encapsulates the model's performance in terms of training accuracy and loss, while Figure 12 delineates the corresponding metrics for testing accuracy and loss. This visual representation affirms the persistent reliability of the identified pattern across the CIFAR-100 dataset.

To better illustrate our model's accuracy metrics differences, we have opted to present our findings using Tables. In Table 1, it is evident that both SwishReLU and ReLU exhibit identical training accuracy of 99.80%. In the testing phase, SwishReLU exhibits slightly superior performance in both accuracy and loss compared to other activation functions.

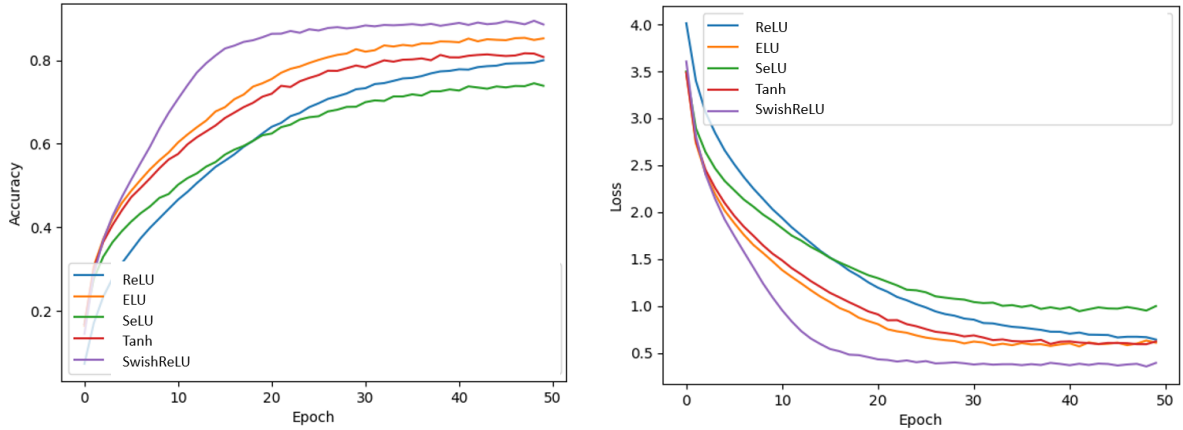

Figure 11: Training data accuracy(left) and loss(right) on the CIFAR-100 dataset.

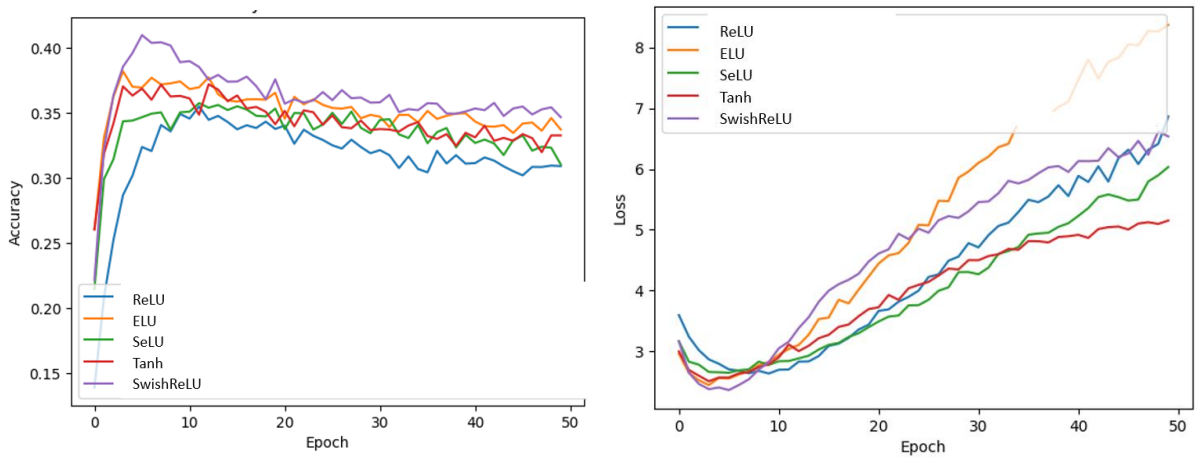

Figure 12: Testing data accuracy(left) and loss(right) on the CIFAR-100 dataset.

Table1: Performance of Multiple activation functions on the MNIST dataset.

|  | Dense Layer | Training Accuracy | Training Loss | Testing Accuracy | Testing Loss |
|---|---|---|---|---|---|
| MNIST | ReLU | 99.80 | 0.0080 | 98.17 | 0.1350 |
|  | ELU | 99.72 | 0.0109 | 98.08 | 0.1693 |
|  | SeLU | 99.61 | 0.0172 | 97.94 | 0.1729 |
|  | Tanh | 99.73 | 0.0080 | 97.66 | 0.1086 |
|  | **SwishReLU** | **99.80** | **0.0080** | **98.23** | **0.1283** |

The comparative analysis presented in Table 2 and Table 3 highlights a notable performance advantage when employing SwishReLU in both the convolutional and dense layers. And it outperforms from others activation functions by 4% and 15% on CIFAR-10 and CIFAR-100 respectively. As depicted in Table 2, the performance metrics for various activation functions on the CIFAR-10 dataset are notable. ReLU achieved a training accuracy of 94% and a testing accuracy of 69%, surpassing SeLU and tanh. ELU demonstrated an improvement with 95% training accuracy and 68% testing accuracy. Notably, SwishReLU stands out as the top performer, surpassing all other activation functions, with an impressive 96% training accuracy and 70% testing accuracy on the CIFAR-10 dataset. In the context of the CIFAR-100 dataset, the findings presented in Table 3 highlight that ReLU achieved a training accuracy of 80%, whereas SwishReLU exhibited remarkable performance, attaining an impressive training accuracy of 88%. Notably, this surpasses the training accuracy achieved by all other activation functions considered in the study, underscoring the effectiveness of SwishReLU in promoting higher accuracy levels during the training phase.

Table 2: Performance of multiple activation functions on CIFAR-10 dataset.

|  | Convolutional Layer | Dense Layer | Training Accuracy | Training Loss | Testing Accuracy | Testing Loss |
|---|---|---|---|---|---|---|
| CIFAR-10 | ReLU | ReLU | 94.73 | 0.1644 | 69.71 | 1.9580 |
|  | ELU | ELU | 95.55 | 0.1640 | 68.79 | 2.8238 |
|  | SeLU | SeLU | 93.61 | 0.2668 | 69.45 | 2.7342 |
|  | Tanh | Tanh | 92.23 | 0.2244 | 67.95 | 1.5584 |
|  | SwishReLU | SwishReLU | **96.11** | **0.1180** | **70.45** | **1.9224** |

Further insights into the dynamic progression of accuracy are provided through visualizations in Figure 13 and Figure 14. These figures present accuracy plots for various activation functions utilized in this study on both datasets CIFAR-10,100. This observation underscores the effectiveness of SwishReLU in achieving higher accuracy levels in both training and testing phases compared to its counterparts in this specific experimental context.

Table 3: Performance of multiple activation functions on CIFAR-100 dataset

| | Convolutional Layer | Dense Layer | Training Accuracy | Training Loss | Testing Accuracy | Testing Loss |
|---|---|---|---|---|---|---|
| CIFAR-100 | ReLU | ReLU | 80.00 | 0.6406 | 30.90 | 6.8633 |
| | ELU | ELU | 85.22 | 0.6073 | 33.71 | 8.3698 |
| | SeLU | SeLU | 73.88 | 0.9973 | 31.03 | 6.0320 |
| | Tanh | Tanh | 80.77 | 0.6214 | 33.26 | 5.1516 |
| | SwishReLU | SwishReLU | **88.55** | **0.3920** | **34.66** | **6.5353** |

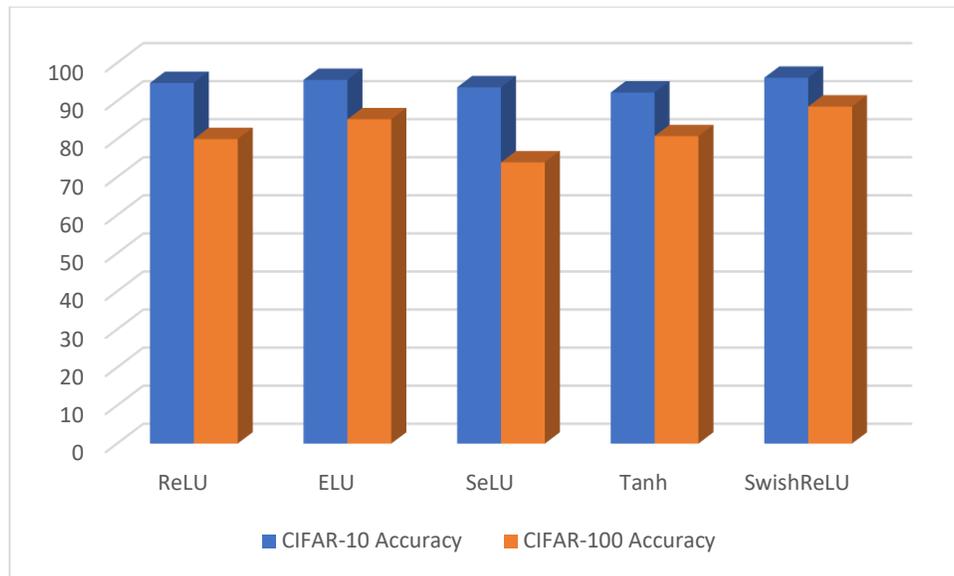

Figure 13: Training Accuracy plot of different Activation functions on the CNN Architecture for CIFAR-10 and CIFAR-100

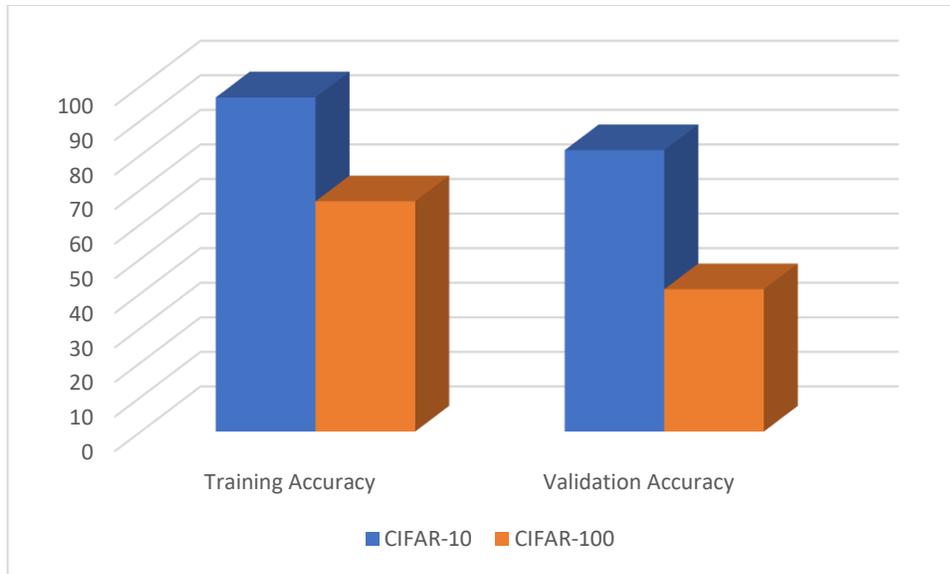

Figure 14: Accuracy plot of VGG-16 Model infused with SwishReLU on CIFAR-10 and CIFAR-100

The observations from both Table 4 and Table 5 clearly indicate that the inclusion of SwishReLU in the dense layers contributes to an enhancement in performance compared to other activation functions. Table 4 provides detailed insights into the performance on the CIFAR-10 dataset. The accuracy of the CNN sees improvement when SwishReLU is integrated into the dense layers. Meanwhile, other activation functions are retained in the convolutional layers. Observing the results, it becomes apparent that there is a consistent 1% rise in both training and testing accuracy for ReLU, ELU, and SeLU. Notably, when SwishReLU is substituted for ReLU in the dense layer, the training accuracy of ReLU experiences an advancement from 94% to 95%. Moreover, there is a corresponding boost in testing accuracy, moving from 69% to 70%. And this observed improvement extends to ELU and SeLU as well. This suggests that SwishReLU, when applied strategically in combination with other activation functions, contributes to enhanced accuracy in the model's performance in comparison to other activation functions on the specified dataset. The consistent improvement extends to the CIFAR-100 dataset, as outlined in Table 5. By initially employing ReLU in the convolutional layers and subsequently replacing ReLU with SwishReLU exclusively in the dense layers, a notable enhancement is evident. The training accuracy of ReLU, which originally stood at 80%, undergoes a substantial increase to 84%. Furthermore, a similar improvement pattern is observed for SeLU and ELU. Specifically, for ELU, the training accuracy experienced an increase from 85% to 87%, while in the case of SeLU, the training accuracy improved from 73% to 79%. This consistent enhancement across multiple activation functions highlights the general effectiveness of incorporating SwishReLU in the dense layers. The utilization of SwishReLU in the dense layers consistently contributes to a superior performance across both datasets, showcasing its effectiveness in enhancing the accuracy of the CNN architecture.

Table 4: Performance of multiple activation functions on CIFAR-10 dataset

| | Convolutional Layer | Dense Layer | Training Accuracy | Training Loss | Testing Accuracy | Testing Loss |
|---|---|---|---|---|---|---|
| CIFAR-10 | ReLU | SwishReLU | 95.45 | 0.1359 | 70.86 | 2.0206 |
| | ELU | SwishReLU | 96.14 | 0.1194 | 70.16 | 1.8856 |
| | SeLU | SwishReLU | 94.95 | 0.1580 | 70.86 | 1.6208 |
| | Tanh | SwishReLU | 92.46 | 0.2186 | 69.15 | 1.4045 |
| | SwishReLU | SwishReLU | 96.11 | 0.1180 | 70.45 | 1.9224 |

Table 5: Performance of multiple activation functions on CIFAR-100 dataset

| | Convolutional Layer | Dense Layer | Training Accuracy | Training Loss | Testing Accuracy | Testing Loss |
|---|---|---|---|---|---|---|
| CIFAR-100 | ReLU | SwishReLU | 84.96 | 0.4979 | 30.58 | 6.7995 |
| | ELU | SwishReLU | 87.52 | 0.4865 | 34.80 | 8.1707 |
| | SeLU | SwishReLU | 79.68 | 0.8403 | 32.64 | 7.5781 |
| | Tanh | SwishReLU | 85.99 | 0.4711 | 33.17 | 6.0862 |
| | SwishReLU | SwishReLU | 88.55 | 0.3920 | 34.66 | 6.5353 |

The VGG16 Model with SwishReLU as an activation function provided 6% accuracy improvement on CIFAR-10 dataset [47]. Applying SwishReLU in the VGG16 model on the CIFAR-10 dataset resulted in an achievement of 96% training accuracy and 81% testing accuracy. Figure 15 shows the training and validation accuracy of the network, illustrating the efficacy of SwishReLU in enhancing the overall performance of the VGG16 model on both datasets. By employing the early stopping criteria, the aforementioned accuracy was attained in a remarkably efficient manner, taking only 11 epochs for CIFAR-10 and 15 epochs for CIFAR-100. Figure 16 illustrates the history of network's loss throughout the training and validation process. This graphical representation clearly indicates that as the number of epochs increases, the loss consistently decreases. Furthermore, observations have been made regarding the feature map illustration from various layers by selecting a random picture for the SwishReLU infused CNN and VGG-16 Model. Figure 17 and Figure 18 serve as visual representations, providing insight into the intricate patterns and representations captured by both models.

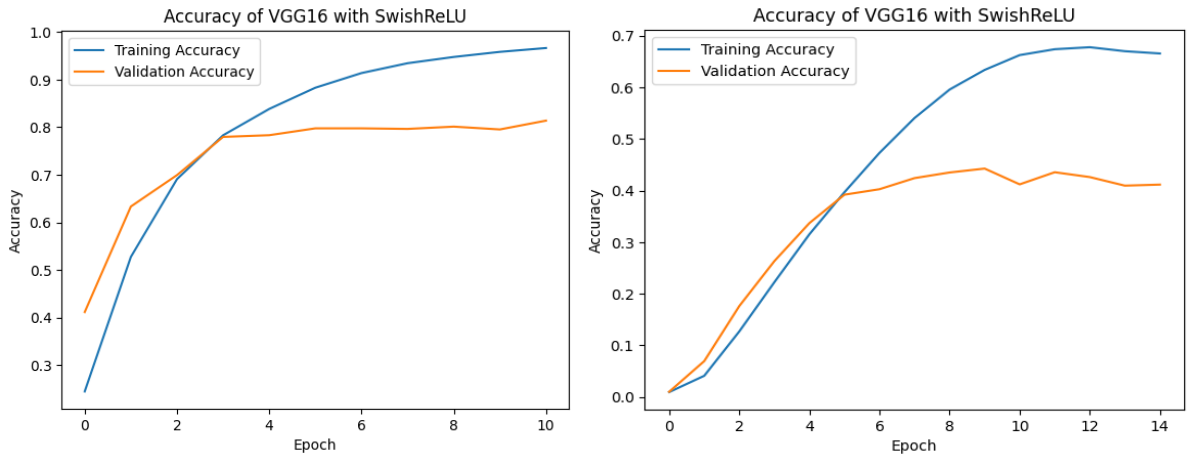

Figure 15: A visual representation depicting the training and validation accuracy of CIFAR-10(left) and CIFAR-100(right)

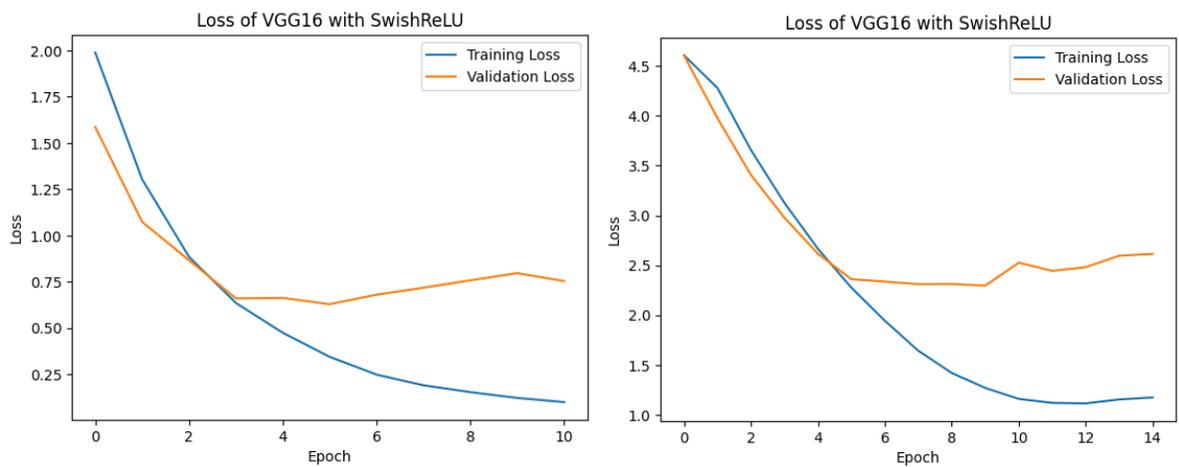

Figure 16: Loss comparison between training and validation sets for CIFAR-10(left) and CIFAR-100(right) when utilizing the SwishReLU activation function.

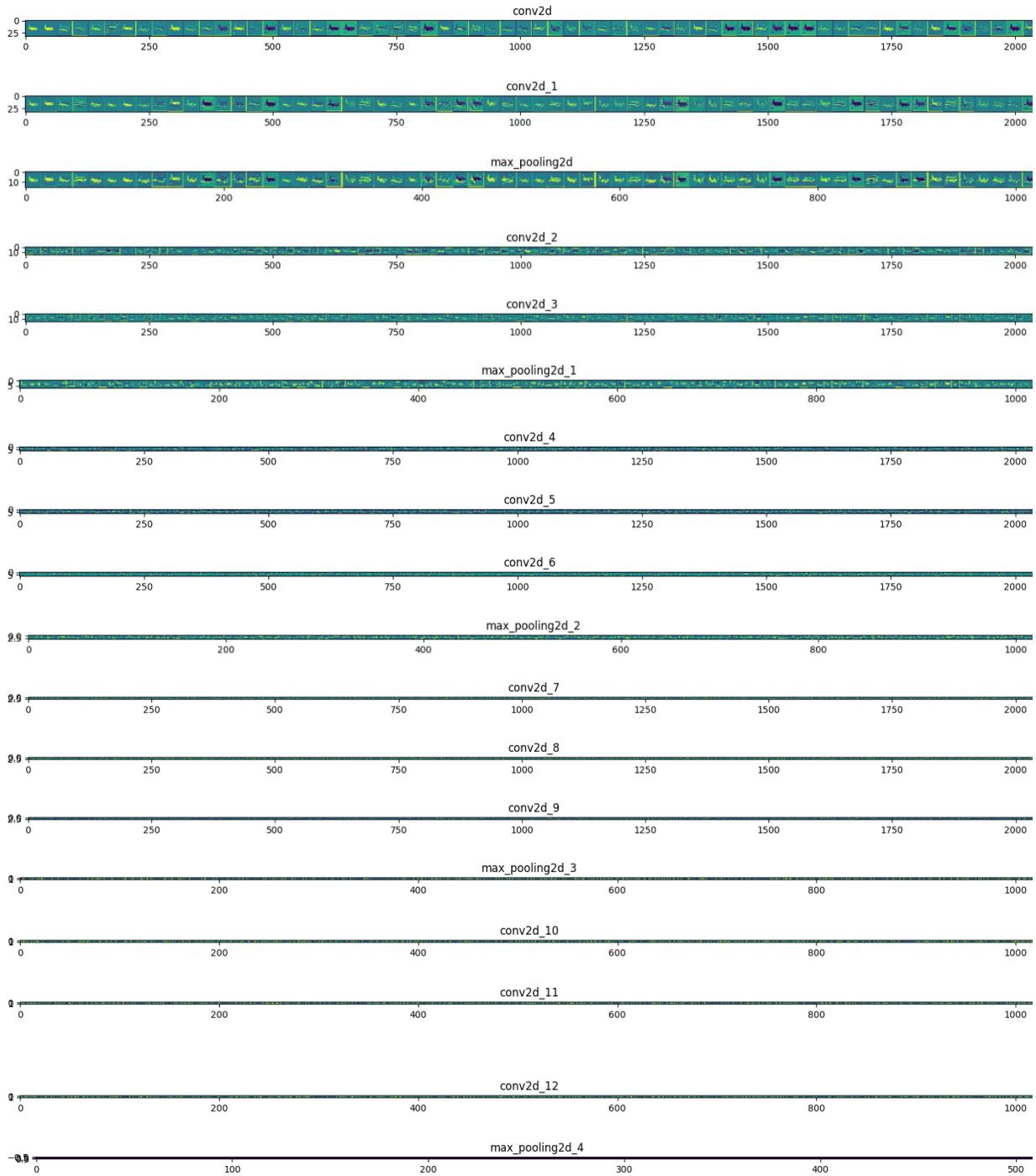

Figure 17: Visualization of feature maps from various layers in the VGG16 model.

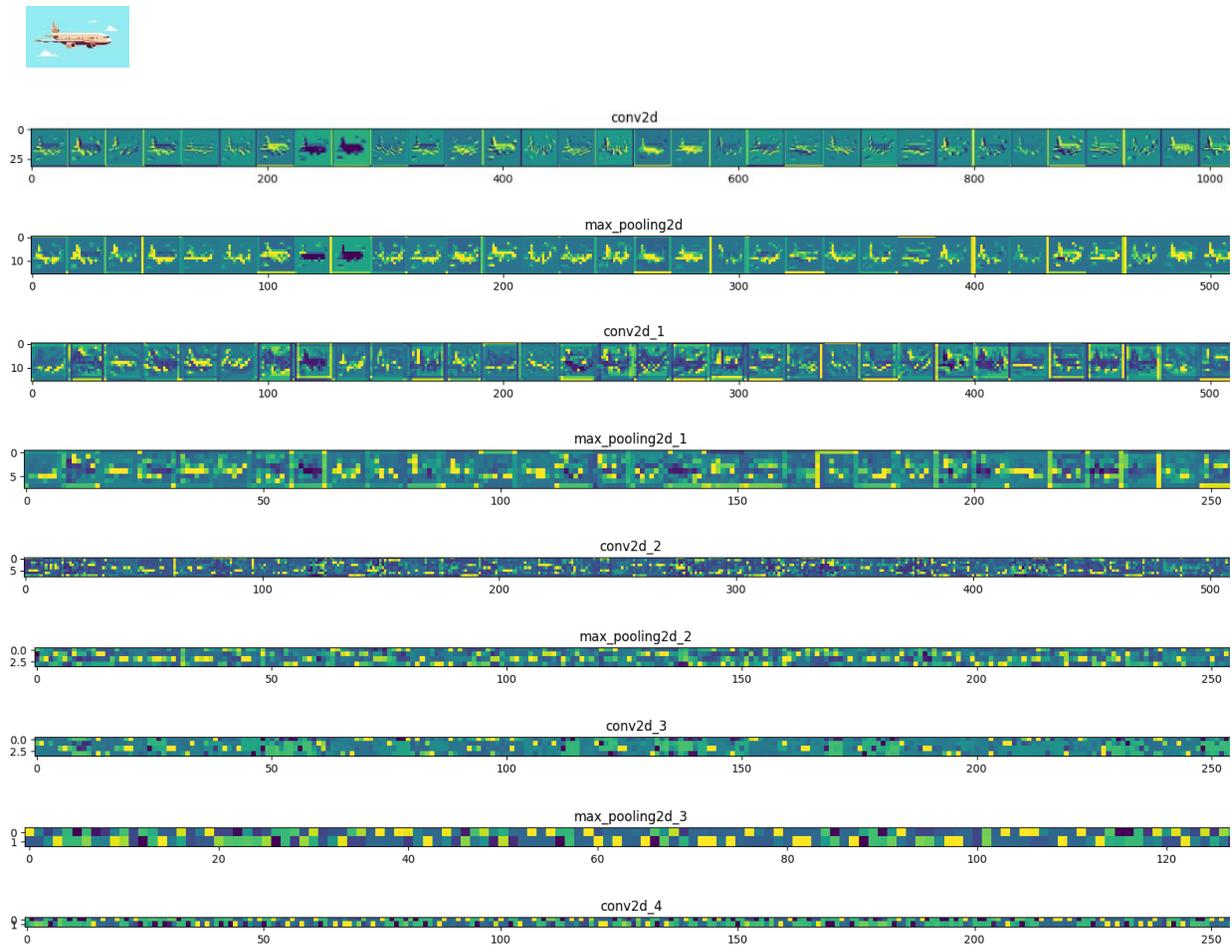

Figure 18: The display of feature maps from several layers of the CNN Architecture

This thorough overview encapsulates key findings and performance metrics, shedding light on the effectiveness of SwishReLU in the realm of image classification. The evaluation spans across various architectures and datasets, providing a comprehensive understanding of how SwishReLU influences and enhances the performance.

Based on the results of our comprehensive evaluation, the proposed methodology demonstrates sufficient strength to be applied with different architectures, loss functions, and optimizers. We are confident that our entire work can be adapted to various versions of the angular softmax loss, as described in related works like CosFace [51], ArcFace [52] and Sphereface [53]. The proposed activation function is differentiable, as described in Section 2. And this characteristic is crucial in Laplacian smoothing gradient descent as it ensures a well-defined gradient, facilitating the application of optimization algorithms [54-55].

## Conclusions:

In this paper, we have proposed a new novel activation function SwishReLU. Based on the experimental results, it has been observed that SwishReLU exhibits notable performance in FCNN for the MNIST dataset, as well as in the CNN and VGG16 models when applied to the CIFAR-10 and CIFAR-100 data. Extensive comparison of performance indicates that SwishReLU outperforms from different activation functions. Additionally, noteworthy improvements in results are observed when switching SwishReLU for other activation functions in dense layers. The VGG16 model with SwishReLU as an activation function has gained 96% accuracy. Feature maps visualization from all convolutional layers of CNN and VGG16 Model further validate the vigilance and effectiveness of the proposed activation. This activation function can be investigated and tested in further computer vision applications in the future.